\documentclass[letterpaper]{article}
\pdfoutput=1

\usepackage{aaai18}
\nocopyright

\usepackage{times}

\usepackage{hyperref}
\usepackage{pgfplots}

\newcommand*{\textcite}[1]{\citeauthor{#1}~\shortcite{#1}}
\newcommand*{\parencite}[1]{\cite{#1}}

\pgfplotsset{
  compat=newest,
  trim axis left,
  trim axis right,
  small,
  height=6.5cm,
  width=8.5cm,
  xlabel={Principal Component 1},
  ylabel={Principal Component 2},
  legend pos=north west,
  legend style={
    font=\footnotesize,
    legend columns=-1,
  },
  scatter/classes={
    a={mark=triangle*, black!33, scale=0.5}
  },
  point meta=explicit symbolic,
}

\title{Learning Taxonomies of Concepts and not Words using\\%
  Contextualized Word Representations: A Position Paper}
\author{
  Lukas Schmelzeisen\textsuperscript{1}
  \and Steffen Staab\textsuperscript{1,\,2}\\
  \textsuperscript{1}Institute for Web Science and Technologies (WeST),
  University of Koblenz-Landau, Germany\\
  \textsuperscript{2}Web and Internet Science Group (WAIS),
  University of Southampton, United Kingdom\\
  \texttt{lukas@uni-koblenz.de},~\texttt{staab@uni-koblenz.de}
}

\begin{document}

\maketitle

\begin{abstract}
  Taxonomies are semantic hierarchies of concepts.
  One limitation of current taxonomy learning systems is that they define
  concepts as single words.
  This position paper argues that contextualized word representations, which
  recently achieved state-of-the-art results on many competitive NLP tasks,
  are a promising method to address this limitation.
  We outline a novel approach for taxonomy learning that%
  ~(1)~defines concepts as synsets,%
  ~(2)~learns density-based approximations of contextualized word
  representations, and%
  ~(3)~can measure similarity and hypernymy among them.
\end{abstract}

\section{Introduction}
\label{sec:intro}

A \emph{taxonomy} is a hierarchical representation of semantic knowledge as a
set of \emph{concepts}~(or \emph{senses}) and a directed acyclic graph of
hyponym--hypernym~(is-a) relations among them.
They have been beneficial in a variety of tasks that require semantic
understanding, such as document clustering~\parencite{DBLP:conf/icdm/HothoSS03},
query understanding~\parencite{DBLP:journals/tkde/HuaWWZZ17},
questions answering~\parencite{DBLP:conf/aaai/YangZWYW17}, and
situation recognition~\parencite{DBLP:conf/cvpr/YatskarOZF17}.
Manually curated taxonomies, such as the one contained in
WordNet~\parencite{DBLP:journals/cacm/Miller95}, are highly precise, but limited
in coverage of both rare concepts and specialized domains.
Therefore, one line of research has focused on creating taxonomies
automatically from natural language corpora, which are widely available for many
domains and languages.

One major limitation of almost all taxonomy learning system is that they
do not distinguish between words and concepts.
In general, however, a many-to-many relationship holds between them.
For example, the word ``Venus'' may either refer to the concept of a
particular planet or to a Roman goddess, and ``morning star'' can either
refer to the same planet or to a type of weapon.
Instead, most automatically constructed taxonomies conflate the different senses
of a word and typically only learn the most predominant one.

This limitation is reinforced by context-free word representations%
\footnote{The NLP literature uses both \emph{distributed representations} and
\emph{distributional representations} to refer to this concept.
The indent is usually to distinguish between latent, dense
vectors~(low-dimensional) and sparse vectors~(high-dimensional), respectively.}%
~\parencite{DBLP:journals/jair/TurneyP10,DBLP:conf/nips/MikolovSCCD13,%
DBLP:conf/emnlp/PenningtonSM14}: methods that encode each word as one point in a
vector space of meaning and are thereby unable to account for multiple senses of
a word.
Such representations are widely used in taxonomy learning and many other
disciplines of natural language processing~(NLP).
In 2018, an emerging trend in NLP have been task-independent deep neural network
architectures based on language model pre-training,
which have achieved state-of-the-art results in a number of competitive
disciplines, such as questions answering or natural language
inference~\parencite{DBLP:conf/naacl/PetersNIGCLZ18,tr/openai/RadfordNSS18,%
DBLP:journals/corr/abs-1810-04805}.
One quality that is common to all of these systems is that they allow for
contextualized word representation: depending on their contexts, occurrences of
the same word can be mapped to the same, similar or very different points in the
vector space.

This position paper argues that contextualized representations, besides being
a powerful input representations for machine learning, provide a promising
approach for distinguishing between the different concepts a word can refer to.
We propose to represent concepts by probability density estimates that
approximate all points in the vector space that correspond to occurrences of the
respective word sense, and to detect whether two concepts are in a taxonomic
relation via divergence between their distributions.
Additionally, our concept representations make it possible to transfer the
strength of contextualized word representations to scenarios where no context
information is available.

The remainder of this paper
reviews the related work in taxonomy learning and word
representation~(Section~\ref{sec:related}),
presents a brief explorative analysis of the lexical semantics encoded by
contextualized word representations~(Section~\ref{sec:analysis}),
details our idea of concept representations for taxonomy
learning~(Section~\ref{sec:approach}), and
concludes~(Section~\ref{sec:conclusion}).

\section{Related Work}
\label{sec:related}

\subsection{Taxonomy Learning}
\label{subsec:taxonomy}

Taxonomy learning typically consists of at least the subtasks term
extraction, hypernym detection, and taxonomy construction%
~\parencite{DBLP:journals/expert/MaedcheS01}.
A recent survey is provided by \textcite{DBLP:conf/emnlp/WangHZ17}.

\subsubsection{Term Extraction}

The goal of this first subtask is to automatically find seed words that are
specific to the domain over which the taxonomy should be constructed.
The domain is usually specified latently via a set of domain-specific
documents.
Most extraction approaches find candidate~\mbox{(multi-)}~words via
part-of-speech patterns and filter them using statistical measures that aim to
estimate theoretical principles such as salience, relevance, or cohesion%
~\parencite{DBLP:conf/iesa/SclanoV07,DBLP:journals/nle/Perinan-Pascual18}.

\subsubsection{Hypernym Detection}

This next task consists of the identification of hypernym--hyponym pairs
over the seed vocabulary~(and possibly additional words).
Pattern-based approaches rely on word pairs occurring in specific
lexicosyntactic patterns in a corpus~\parencite{DBLP:conf/acl/RollerKN18}.
They achieve high precision but suffer from low recall because they rely on both
words occurring together in context.
A second type of approach is based on vector representations of words, and is
thus not reliant on local context.
In the unsupervised case, usually a score is assigned to each possible word
pair, which is expected to be higher for hypernym pairs than for negative
instances.
Many competing methods exist of which none is clearly superior to others;
they differ both in the construction of the vector space as well as the employed
detection measure~\parencite{DBLP:conf/eacl/SantusSS17,%
DBLP:conf/naacl/ChangWVM18}.
Supervised approaches typically achieve better results on existing benchmarks,
but have been shown to not truly detect hypernymy, but rather which words are
prototypical hypernyms, making them unreliable for real-world
applications~\parencite{DBLP:conf/naacl/LevyRBD15,DBLP:conf/eacl/SantusSS17}.
Pattern- and vector-based approaches have been integrated successfully
by \textcite{DBLP:conf/acl/ShwartzGD16} using LSTM neural networks.

\subsubsection{Taxonomy Construction}

In this subtask the final taxonomic hierarchy is constructed from identified
hypernym pairs.
This is non-trivial because identified hypernym pairs are noisy and basically
never induce a connected directed acyclic graph.
Approaches can be divided into clustering-based~\parencite{%
DBLP:journals/dke/KnijffFH13} and graph-based ones~\parencite{%
DBLP:journals/coling/VelardiFN13,DBLP:conf/cikm/GuptaLHA17}.
Afterwards, clean-up operations like cycle elimination are sometimes
performed~\parencite{DBLP:conf/aaai/LiangXZHW17}.

\subsubsection{Discussion}

To the best of our knowledge, there is no taxonomy learning system providing
a principled way to model multiple word senses.
At most, existing system consider in which grammatical role a word occurs
in~\parencite{DBLP:conf/eacl/SantusSS17}, which is often a bad signal for
separating word senses, or explicitly try to word sense
disambiguate the input corpus in a preprocessing
stage~\parencite{DBLP:conf/naacl/KlapaftisM10}, which is often error-prone.
As a result, automatically created taxonomies conflate word senses and
important semantic properties like transitivity of the hypernymy relation
do not usually hold~\parencite{DBLP:conf/aaai/LiangZXW0Z17}.

\subsection{Word Representations}
\label{subsec:representation}

\subsubsection{Context-free Word Representations}

Most \emph{context-free word representations} map each word to a single point in
an often latent vector space of meaning---regardless of the context the word is
used in---with the goal to place words with similar meaning close to each
other~\parencite{DBLP:journals/jair/TurneyP10}.
After groundbreaking results achieved by factorizing word co-occurrence
matrices~\parencite{DBLP:conf/nips/MikolovSCCD13,%
DBLP:conf/emnlp/PenningtonSM14},
such representations have been used to encode words in most state-of-the-art NLP
architectures since.
These representations are only able to represent one sense per word as they map
each word to exactly one vector.
One strand of research has thus focused on mapping each word to multiple
vectors that should each represent a different word sense%
~\parencite{DBLP:journals/corr/abs-1805-04032}.
The number of senses per word have been either learned by clustering contexts,
in a preceding step~\parencite{DBLP:conf/acl/HuangSMN12} and jointly during
model training~\parencite{DBLP:conf/emnlp/NeelakantanSPM14}, or specified via a
given sense inventory~\parencite{DBLP:conf/emnlp/ChenLS14}.
A related line of work criticizes the assumption that all meanings of a word can
be discretely separated, and instead represent each word as probability
densities in the vector space---usually as Gaussian distributions%
~\parencite{conf/iclr/VilnisM15,DBLP:conf/acl/AthiwaratkunW17}.
The general meaning of a word is then characterized by the mean vector of the
distribution analogous to before, while additionally the covariance can be
interpreted as spread of meaning or uncertainty.
We still classify these advanced representation approaches as context-free,
because even though their main training signal is what contexts a word occurs
in, they are unable to adapt or select the representation for a word given a
specific context.
Possibly as a consequence of this, they have seen almost no adoption in
practice.
Specifically, we are not aware of any taxonomy learning approaches building on
these representations.

\subsubsection{Contextualized Word Representations}

Until very recently, almost all state-of-the-art solutions in NLP were highly
specialized task-specific architectures.
In contrast, in 2018 three related task-agnostic architectures have been
published that achieved state-of-the-art results across a wide range of
competitive tasks, suggesting some generalizable language understanding for the
first time.
All of these systems built on the language modeling objective: training a
model to predict a word given its surrounding context.
Because such training examples can be built from unlabeled corpora, much
training data is available.
ELMo~\parencite{DBLP:conf/naacl/PetersNIGCLZ18} trains representations with
stacked bidirectional LSTMs, but still employs task-specific architectures on
top of them.
OpenAI GPT~\parencite{tr/openai/RadfordNSS18} and BERT~\parencite{%
DBLP:journals/corr/abs-1810-04805} do away with this and instead train
task-agnostic transformer stacks that are only fine-tuned together with a single
dense layer for each downstream task.
The latter mainly improves upon the former by joint conditioning on both
preceding and following contexts.
Critically, all systems allow for \emph{contextualized word representation}:
they map each word occurrence to a vector specifically considering the
surrounding context.
Much of their success is attributed to the ability to better
disambiguate polysemous words in a given sentence.
This representation approach is easily applicable for many NLP tasks, where
inputs are usually sentences and context information is thus available.
However, due to the hierarchical nature of taxonomies, there is no
straightforward way to utilize the generalization power of contextualized
representations for our task.

\begin{figure*}
  \centering
  \hspace{1.25cm}
  \begin{tikzpicture}
    \begin{axis}[
      enlargelimits=0.15,
    ]
      \addplot [
        scatter, mark=x, only marks,
        nodes near coords,
        nodes near coords style={
          font=\scriptsize,
          xshift=\xShift,
          yshift=\yShift,
        },
        visualization depends on={\thisrow{xshift} \as \xShift},
        visualization depends on={\thisrow{yshift} \as \yShift},
      ] table[meta=label] {
        x      y      label xshift yshift
        -3.4   -0.01  town -0.15cm -0.05cm
        -3.34  -0.05  province 0 -0.35cm
        -3.20   0.29  country 0 -0.05cm
        -2.57  -0.52  election 0 -0.05cm
        -2.61  -0.69  survey 0 -0.4cm
        -2.65  -0.65  campaign -0.55cm -0.23cm
        -2.37  -2.17  speech 0 -0.05cm
        -2.34  -2.37  sentence 0.55cm -0.18cm
        -2.37  -2.50  announcement 0 -0.4cm
        -2.65  -2.40  message -0.55cm -0.23cm
        -0.55   4.17  national -0.05cm 0
        -0.21   4.61  federal 0 0
         0.47   4.85  domestic 0 0
         0.14   4.27  international 0 -0.4cm
        -1.49   4.12  military -0.2cm -0.05cm
        -1.28   3.81  terrorist 0 -0.4cm
        -1.02   3.97  rebel -0.1cm 0
        -0.54   0.30  earnings 0.55cm -0.23cm
        -0.70   0.11  savings 0 -0.45cm
        -0.89   0.40  income -0.5cm -0.18cm
        -0.80   0.51  funding 0 -0.05cm
         0.35   1.61  Bush -0.1cm 0
         0.78   1.54  Clinton 0.1cm 0
         0.71   1.40  Obama 0 -0.4cm
        -0.07  -1.94  speculation -0.7cm -0.23cm
         0.34  -1.72  forecast 0 0
         0.06  -2.07  estimate 0 -0.4cm
        -2.27   1.93  governor 0.2cm -0.05cm
        -2.89   2.02  ministry 0 -0.05cm
        -3.04   1.83  embassy 0 -0.4cm
        -2.13   1.58  president 0 -0.45cm
      };
    \end{axis}
  \end{tikzpicture}
  \hspace{0.82cm}
  \begin{tikzpicture}
    \begin{axis}[
      yticklabel pos=right,
      enlargelimits=0.05,
      legend entries={
        animal
      },
    ]
      \addplot [
        scatter, only marks,
      ] table [meta=label] {
        x      y      label
         0.63   5.60  a
        -1.04   6.88  a
        -2.24   2.73  a
        -3.40   0.57  a
        -0.14   3.99  a
         0.53   5.44  a
        -0.54   5.52  a
        -2.56   4.50  a
         0.07   4.70  a
         0.21   4.34  a
         0.47   5.26  a
        -3.72   0.54  a
         0.31   4.93  a
        -0.32   5.59  a
         0.02   6.53  a
        -3.85   2.33  a
        -3.28   0.08  a
        -2.04   2.38  a
        -2.08   2.27  a
        -0.18   5.47  a
        -0.53   5.44  a
        -0.08   4.64  a
         0.62   5.15  a
        -0.92   4.64  a
        -3.48   2.27  a
         0.05   6.47  a
        -1.90   3.88  a
        -3.41  -0.90  a
        -2.82   3.87  a
        -3.74   2.69  a
        -1.71   1.77  a
         0.46   5.56  a
         0.39   4.67  a
        -3.26   4.05  a
        -2.93   2.90  a
        -0.52   4.31  a
        -2.25   3.49  a
        -3.96   2.72  a
        -3.10   3.16  a
        -3.23   0.32  a
        -1.61   5.08  a
        -2.19   3.85  a
         0.24   4.70  a
         0.77   4.99  a
        -3.24   0.05  a
        -2.10   0.59  a
         0.44   5.99  a
        -0.00   5.72  a
         0.05   4.54  a
        -1.04   5.27  a
        -3.62   0.76  a
        -2.45   2.55  a
        -4.00   3.17  a
        -0.78   4.44  a
        -3.72   3.26  a
        -0.26   4.86  a
        -2.77   0.20  a
        -0.33   5.30  a
         0.13   4.82  a
        -2.54   4.77  a
        -0.01   4.40  a
         0.24   3.35  a
        -3.26   0.95  a
        -0.27   4.97  a
        -2.25   2.37  a
        -3.16   4.45  a
         0.38   5.04  a
         0.09   4.20  a
        -2.89   2.47  a
        -1.05   4.48  a
        -3.36   0.05  a
        -3.71   1.26  a
        -2.83   2.45  a
        -3.12   0.13  a
        -2.52   3.10  a
        -2.86   2.71  a
        -0.15   7.27  a
        -2.68   3.56  a
        -3.42   0.87  a
        -1.83   2.50  a
        -0.27   3.73  a
        -0.36   4.26  a
        -0.27   2.65  a
        -1.18   4.14  a
        -0.13   5.92  a
        -2.72  -1.16  a
        -2.86   2.89  a
        -0.04   5.56  a
        -1.16   4.55  a
        -3.20   1.20  a
        -2.95   1.82  a
        -2.68   1.84  a
        -0.14   4.82  a
        -0.42   5.18  a
        -3.63   1.82  a
        -3.83   3.33  a
        -3.36  -0.31  a
        -0.84   3.39  a
        -3.73   2.39  a
        -1.93   3.30  a
        -2.84   1.25  a
        -1.73   2.45  a
        -3.51  -0.54  a
         0.22   5.36  a
        -3.69   2.66  a
        -2.58  -0.09  a
         0.16   5.79  a
        -3.48   1.13  a
         0.31   4.72  a
        -3.43   2.51  a
        -2.70   1.83  a
        -3.16   3.55  a
        -2.03   5.29  a
        -2.46   1.74  a
        -2.71  -0.11  a
        -3.81   2.01  a
        -1.59   1.67  a
         0.28   4.52  a
        -0.89   4.89  a
         0.25   4.51  a
         0.77   4.50  a
        -3.72   2.16  a
        -0.88   6.16  a
        -2.90   1.44  a
        -3.14   0.11  a
        -3.53  -0.15  a
         0.46   5.02  a
        -1.51   1.95  a
        -0.54   3.74  a
        -3.38   2.83  a
        -0.03   5.17  a
        -0.78   1.37  a
        -0.87   3.84  a
        -3.06   1.67  a
        -0.21   4.17  a
        -3.26  -0.68  a
        -2.93  -0.28  a
        -0.97   5.42  a
        -0.37   6.31  a
        -2.24   4.15  a
        -0.60   4.55  a
        -3.94   0.45  a
        -2.05   4.17  a
        -0.41   3.32  a
        -1.10   4.35  a
         0.26   5.84  a
        -1.85   5.02  a
        -0.12   3.99  a
        -3.32   3.57  a
        -0.94   3.82  a
        -0.12   4.77  a
        -1.95   4.10  a
        -2.58   0.86  a
        -3.39  -0.03  a
        -2.98  -0.30  a
        -2.05   0.28  a
        -0.49   5.17  a
        -0.68   3.79  a
        -3.80   3.50  a
        -3.26   1.50  a
        -2.21   1.76  a
         0.21   4.92  a
        -0.43   4.27  a
        -2.82   3.29  a
        -0.13   5.41  a
        -3.11   0.55  a
        -0.13   6.12  a
        -0.36   3.86  a
        -0.19   4.98  a
         0.10   3.06  a
        -0.09   5.22  a
         0.33   5.60  a
        -0.21   4.58  a
        -2.82   2.88  a
        -3.09  -0.94  a
        -2.21   2.15  a
         0.24   4.87  a
        -0.26   6.16  a
        -2.54   3.24  a
        -0.40   4.51  a
        -0.08   4.05  a
        -1.79   4.38  a
        -3.16   2.88  a
        -2.89   5.10  a
        -2.54  -0.95  a
        -2.46   1.67  a
        -3.27   0.99  a
        -2.99  -0.25  a
        -0.47   3.99  a
         0.20   5.30  a
        -3.71   3.15  a
        -0.02   5.63  a
        -1.83   5.29  a
        -0.21   4.67  a
         0.84   5.06  a
        -0.48   6.13  a
        -0.22   5.44  a
        -0.70   2.48  a
        -2.07  -1.08  a
        -1.11   5.36  a
        -3.25   3.03  a
        -3.29   3.72  a
        -2.44  -0.45  a
        -1.02   3.72  a
        -3.34   1.41  a
        -0.96   5.26  a
        -2.97   5.26  a
        -0.42   5.21  a
        -3.44   0.51  a
        -0.26   4.41  a
        -3.43   2.75  a
         0.59   5.51  a
        -3.93   2.64  a
        -0.43   5.44  a
        -3.12  -0.73  a
        -2.24   3.35  a
        -3.48   2.59  a
         0.25   6.03  a
        -2.77   0.19  a
        -2.76  -0.21  a
        -0.95   4.18  a
         0.25   5.40  a
         0.24   4.99  a
        -3.24   1.34  a
        -3.34   2.27  a
        -0.29   4.91  a
        -0.29   5.65  a
         0.35   5.33  a
        -3.45   0.51  a
        -2.90   0.91  a
        -3.40  -1.05  a
         0.11   4.36  a
        -0.08   5.28  a
         0.32   4.53  a
        -3.15   1.10  a
        -0.01   4.09  a
        -3.52  -0.10  a
        -4.41   0.93  a
        -2.88  -0.92  a
        -0.04   6.96  a
        -3.40   2.11  a
         0.06   5.81  a
         0.27   4.62  a
        -2.16   3.01  a
        -3.40   2.46  a
        -3.02   4.00  a
        -2.41   4.84  a
        -0.09   4.39  a
         0.12   6.68  a
         0.38   4.89  a
        -3.20  -0.66  a
        -0.23   5.70  a
        -0.63   3.43  a
        -2.24   3.61  a
        -3.64   3.22  a
         0.18   3.96  a
        -0.10   4.06  a
        -3.36   2.85  a
        -2.28   1.32  a
        -0.77   7.77  a
        -0.14   3.87  a
        -3.28  -0.48  a
        -3.02  -0.50  a
        -0.33   3.88  a
        -0.44   5.88  a
        -0.18   6.37  a
        -0.69   5.68  a
        -3.66   3.17  a
        -0.81   4.42  a
        -3.06   0.01  a
        -2.14   3.45  a
        -2.98   0.09  a
         0.08   6.17  a
        -2.37   1.32  a
        -0.08   5.71  a
        -2.24   3.17  a
        -2.70   4.31  a
        -0.05   5.07  a
        -3.97   1.98  a
        -0.71   3.68  a
        -3.26   2.80  a
        -1.83   2.30  a
        -2.68   0.46  a
        -3.45   0.49  a
         0.36   5.21  a
         0.26   4.75  a
        -2.56   1.15  a
        -3.31   0.08  a
        -0.16   5.44  a
        -0.71   5.19  a
        -0.22   5.47  a
        -3.19   0.74  a
        -2.65   2.88  a
        -0.22   4.03  a
         0.68   4.74  a
        -1.11   0.89  a
         0.15   5.36  a
        -0.57   4.53  a
        -3.32   2.04  a
        -3.52   1.82  a
        -0.11   5.25  a
        -2.16   1.15  a
        -2.96   0.75  a
        -2.75  -0.41  a
        -0.03   4.36  a
         0.12   4.66  a
      };

      \addplot [
        densely dotted,
        very thick,
        black!33,
        mark=triangle,
        every mark/.append style={solid},
      ] table {
        x      y
        -0.77   7.77
        -2.97   5.26
        -4.00   3.17
        -4.41   0.93
        -3.40  -1.05
        -2.72  -1.16
        -2.07  -1.08
        -0.78   1.37
         0.10   3.06
         0.24   3.35
         0.77   4.50
         0.84   5.06
         0.63   5.60
        -0.15   7.27
        -0.77   7.77
      };

      \addplot [
        scatter, mark=x, only marks,
        nodes near coords,
        nodes near coords style={
          font=\scriptsize,
          xshift=\xShift,
          yshift=\yShift,
        },
        visualization depends on={\thisrow{xshift} \as \xShift},
        visualization depends on={\thisrow{yshift} \as \yShift}
      ] table [meta=label] {
        x      y      label xshift yshift
        -2.02   3.13  alligator 0 -0.05cm
        -1.73   2.35  bee 0 -0.05cm
        -2.13   2.63  bird -0.25cm -0.05cm
        -2.07   1.37  cat 0.2cm -0.07cm
        -1.16   2.42  chicken 0.2cm 0
        -1.82   1.17  cow 0 -0.35cm
        -2.11   1.50  dog -0.1cm -0.05cm
        -2.61   2.10  elephant 0 -0.05cm
        -2.70   1.40  frog -0.3cm -0.23cm
        -2.41   1.16  horse 0 -0.35cm
        -2.56   1.81  lion -0.3cm -0.17cm
        -1.16   2.22  pig 0.3cm -0.21cm
        -1.12   1.99  sheep 0.18cm -0.42cm
        -2.07   0.58  snake 0 -0.35cm
        -3.49   2.35  tiger -0.3cm -0.23cm
        -1.54   1.29  wolf 0.15cm -0.35cm
      };
    \end{axis}
  \end{tikzpicture}
  \hspace{1.25cm}
  \caption{PCA projections of 1024-dimensional ELMo vectors~(all biLM layers
    averaged).
    Shown words were selected manually.
    X's indicate the projection of the centroid of all representations for their
    respective label.
    Additionally, in the right panel, triangles are the projections of all
    representations of the word ``animal'' together with their convex
    hull as a dotted line.}
  \label{fig:visual}
\end{figure*}
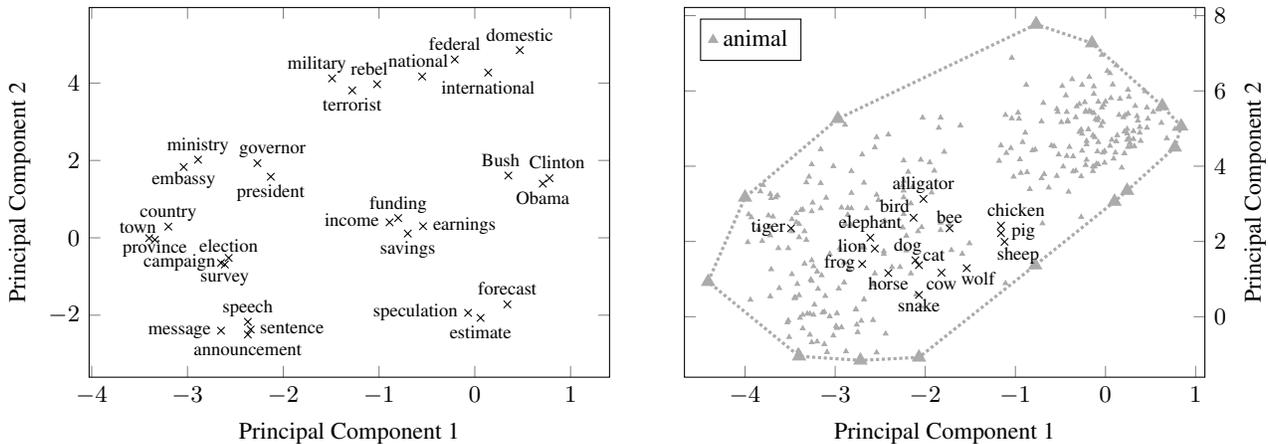

\section{Explorative Analysis}
\label{sec:analysis}

We present a brief explorative analysis of the lexical semantics encoded by
contextualized word representations here, to prepare the argument in
Section~\ref{sec:approach}.

Figure~\ref{fig:visual} shows two visualizations of ELMo representations.
For this, the pre-trained model \texttt{elmo\_2x4096\_512\_\allowbreak{}%
2048cnn\_2xhighway}\footnote{Available at: \url{https://allennlp.org/elmo}} was
used to compute all contextualized word representations on the first percent
(roughly 200\,000 different words and 7\,500\,000 word occurrences) of the
corpus by \textcite{DBLP:conf/interspeech/ChelbaMSGBKR14}, a corpus of news
paper articles on which the model was originally trained on by
\textcite{DBLP:conf/naacl/PetersNIGCLZ18}.

We observe a number of interesting properties:
\begin{itemize}
  \item The centroids of all ELMo representations of a word tend to show
    strong local clustering behavior based on word similarity.
    Specifically, in the left panel of Figure~\ref{fig:visual} this is visible
    for a number of words from the political domain.
    Such behavior is not immediately relevant to taxonomy learning, but since
    encoding similarity is the main goal of context-free word
    representations~\parencite{DBLP:journals/jair/TurneyP10} this suggest
    a strong representational power for ELMo, that is preserved when only
    considering centroids.
    Clustering was even better visible when visualizing with
    t-SNE~\parencite{journal/jmlr/MaatenH08}, but we only show PCA projections
    here for consistently with the next point.
  \item The representations of hypernym words tend to exhibit a larger spread in
    the vector space than the representations of their hyponyms.
    This motivates our approach for detecting hypernymy using contextualized
    representations, which we detail in the next section.
    Specifically, we visualize this behavior in the right panel of
    Figure~\ref{fig:visual}, where we show that the centroids of words for
    different animal species lie inside the convex hull of the representations
    of ``animal''.
    Surprisingly, this behavior was not visible when visualizing with t-SNE.
    We suspect that this it the case because of an implementation detail of
    ELMo, namely the final contextualized representations being derived from
    context-free ones, and that thereby the contextualized representations are
    still separable in high-dimensional space, which is the optimization
    objective of t-SNE.
  \item For words with multiple senses, the representations of individual senses
    tend to lie close to clusters of other words that are similar in meaning
    to the respective sense.
    This suggests their suitability for discovering synsets of words, as we
    propose in the next section.
    Because of limited space, we do not show this behavior here, but
    \textcite{DBLP:conf/emnlp/StanovskyH18} made similar observations.
\end{itemize}

\section{Concept Representations}
\label{sec:approach}

In this section, we outline our idea for concepts representations from
contextualized word representations.
Their aim is to address the current limitations of current taxonomy learning
systems, namely that they do not distinguish between words and concepts,
which we believe to be a major hindering factor for learning accurate
taxonomies.
Our approach mainly consists of changes in the subtasks of term extraction and
hypernymy detection whereas for constructing the final taxonomies existing
techniques can be used without major modification.

For term extraction, our idea reuses existing work for finding domain-specific
seed words by estimating the relevance of words to the target
domain~\parencite{DBLP:journals/nle/Perinan-Pascual18}.
Going beyond previous work though, we propose to learn synsets of word senses
for defining concepts instead of just defining them by single words.
A \emph{synset} is a set of different words each sharing a common sense
interpretation.
They have been popularized by WordNet~\parencite{DBLP:journals/cacm/Miller95},
and are a psychologically plausible definition for concepts, as humans are
usually able to infer the specific sense meant as the intersection of all words
in a synset~\parencite{DBLP:conf/emnlp/StanovskyH18}.
For finding synsets, we propose to%
~(1)~calculate the contextualized word representation vectors for all word
occurrences in the given corpus,%
~(2)~group vectors using a clustering algorithm, and, for retaining only
domain-specific concepts,
to optionally~(3)~filter out all clusters not containing at least one seed word
as previously determined via term extraction.
Each resulting cluster of word vectors then constitutes one synset.
We leave open what clustering algorithm should be chosen for this, although
simple ones like $k$-means clustering might already be sufficient.
Here, the parameter $k$ would control into how many concepts the vector space
should be separated.
Indeed, $k$-means clustering of ELMo representations has been shown to work well
by \textcite{DBLP:conf/emnlp/StanovskyH18}.
This is also the only work in this direction that we are aware of: the authors
cluster all ELMo vectors belonging to a specific word to determine the number of
senses of that word, but do not investigate properties of clustering all ELMo
vectors of a corpus.
Alternatively, to avoid choosing a parameter $k$ Dirichlet processes could be
used.

The result of the previous step is a set of clusters (of contextualized word
representation vectors) each characterizing one synset/concept.
For determining similarity and hypernymy among these concepts as well as for a
more parameter efficient representation, we propose to learn what we call
\emph{concept representations}: probability density estimates that approximate
all vectors belonging to one concept.
In continuation of current research~\parencite{conf/iclr/VilnisM15,%
DBLP:conf/acl/AthiwaratkunW17}, we specifically suggest to use Gaussian
distributions although others are conceivable, such as Student's
$t$-distribution.
For Gaussians, the mean vector is usually interpreted as characterizing the
general meaning of a word whereas the covariance signifies the
generality/unspecificity of the word.
Following \textcite{conf/iclr/VilnisM15}, the mean and covariance of the density
can be found by averaging all vectors of a synset or calculating the empirical
covariance among them, respectively.
\textcite{DBLP:conf/acl/AthiwaratkunW17} suggest to learn multimodal Gaussian
distributions, but we deem this unnecessary in our case, since the input vectors
for each distribution should already be comparatively close together as a result
of the preceding clustering step.

For context-free representations, similarity of words is usually estimated using
the dot product of the respective vectors.
In an analogous way, similarity for density-based representations can be
estimated via the inner product (probability product kernel) of the respective
densities~\parencite{conf/iclr/VilnisM15}.
Our method for determining hypernymy is motivated by the
\emph{distributional inclusion hypothesis}~\parencite{DBLP:conf/acl/GeffetD05},
which states that a hypernym--hyponym relation holds among word senses exactly
when the hypernym sense can occur in all the contexts the hyponym sense can
occur in.
Since contextualized word representations characterize exactly which semantic
context holds for a given word occurrence, our concept representations should
characterize which semantic contexts a concept can occur in.
There, we formulate the problem of detecting hypernymy among two concepts as
to what degree one concept's density is ``included'' in the other one's.
\textcite{conf/iclr/AthiwaratkunW18} survey a number of measures for this, the
most well-known being the Kullback-Leibler divergence.
One advantage of this approach over conventional ones is that it naturally
allows to characterize the strength of the hypernymy relation, as motivated by
\textcite{DBLP:journals/coling/VulicGKHK17}.

\subsection{Caveats}

Our approach requires computing and storing the contextualized word
representations of all word occurrences for a training corpus with a given
pre-trained model, which is already resource intensive%
\footnote{Specifically, we measured that calculating ELMo representation with
the pre-trained model took about 5\,ms per word token~(roughly 58\,days per
billion tokens) on an AMD Ryzen Threadripper 1950X CPU and about 0.2\,ms per
token~(2.5\,days per billion tokens) on a NVIDIA Titan~V GPU.
Storing the calculated representations took about 8\,kB per token~(8\,TB per
billion tokens).}.
Training a new model from scratch is orders of magnitude more expensive.

Additionally, being able to handle multiword expressions~(MWEs), such as
``morning star'', is of critical importance to taxonomy learning techniques,
since most non-trivial concepts do not have single word identifiers.
There has been no research so far on finding a single vector encoding for a
given MWE from the contextualized word representation of its constituent words,
which is hard because the semantics of most MWEs are non-compositional.
However, this is not a fundamental problem to our approach since MWEs can just
be mapped to single words, such as ``morning\_star'' in a preprocessing step
before training the contextualized word representations.
The same technique is commonly performed for
word2vec~\parencite{DBLP:conf/nips/MikolovSCCD13}.
Though this means, that no pre-trained model could be used.

\section{Conclusion}
\label{sec:conclusion}

We have sketched an approach for learning taxonomies using contextualized
word representations that distinguishes words and concepts.
Our outlined idea is novel in that we are the first to propose%
~(1)~defining concepts in automatically constructed taxonomies as synsets, which
we plan to learn by clustering contextualized word representations,%
~(2)~approximating a set of related contextualized word representations via
probability density estimates, which we call concept representations, and
~(3)~using such representations to determine similarity and hypernymy among
concepts.

Further, our concept representations provide the opportunity to study other
interesting semantic relations among concepts, for example, modeling union and
intersection of concepts via the union and intersection of their densities,
respectively.
Additionally, they could find usage in a number of NLP tasks as substitutes for
the contextualized word representations that they approximate, which take
comparatively long times to compute even on powerful GPUs, since our
representations are far more parameter efficient.

Beyond our proposed idea of using contextualized word representations for
taxonomy learning, which is in continuation of existing distributional hypernymy
detection techniques, we foresee contextualized word representations also being
used to advance pattern-based approaches. Here, they would allow to learn not
only lexicosyntactic but, for the first time, also semantic patterns that
indicate hypernymy.

\hbadness=10000 
\small
\bibliographystyle{aaai}
\bibliography{schmelzeisen.bib}

\end{document}